\documentclass[letterpaper,journal]{IEEEtran}
\usepackage{amsmath,amsfonts,amssymb}
\usepackage{algorithmic}
\usepackage{algorithm}
\usepackage{array}
\usepackage[caption=false,font=normalsize,labelfont=sf,textfont=sf]{subfig}
\usepackage{textcomp}
\usepackage{stfloats}
\usepackage{url}
\usepackage{verbatim}
\usepackage{graphicx}
\usepackage{adjustbox}
\usepackage{multirow}
\usepackage{cite}
\usepackage{booktabs}
\usepackage{tabularx}
\usepackage{wrapfig}
\usepackage{amsthm}
\usepackage{xcolor}
\usepackage{pifont}

\usepackage{orcidlink}
\newcommand{\Checkmark}{\ding{51}}
\newcommand{\XSolid}{\ding{55}}

\begin{document}

\title{PhysFlow: Physics-Aware Optical Flow for \\Motion Controllable Video Generation}

\author{
Cong Wang~\orcidlink{0009-0007-9702-7953}, Hanxin Zhu~\orcidlink{0009-0006-3524-0364}, Yonglin Tian~\orcidlink{0000-0003-1911-5791}, Jiayi Luo~\orcidlink{0009-0004-5742-3589}, Ruiqi Song~\orcidlink{0000-0003-2261-3724}, Boyi Sun~\orcidlink{0009-0005-6780-7495}, \\Long Chen~\orcidlink{0000-0003-4925-0572},~\IEEEmembership{Senior Member,~IEEE}, and Zhibo Chen~\orcidlink{0000-0002-8525-5066},~\IEEEmembership{Senior Member,~IEEE}
\thanks{\textcolor{black}{This work was supported by the Zhongguancun Academy under Project C20250302. (\textit{Cong Wang and Hanxin Zhu contributed equally to this work}). (\textit{Corresponding author: Long Chen, Zhibo Chen}).}}
\thanks{
Cong Wang is with the State Key Laboratory of Multi-modal Artificial Intelligence Systems, Institute of Automation, Chinese Academy of Sciences, Beijing 100190, China, with the School of Artificial Intelligence, University of Chinese Academy of Sciences, Beijing, 100049, China, and also with Zhongguancun Academy, Beijing 100094, China (e-mail: wangcong2024@ia.ac.cn).
}
\thanks{
Hanxin Zhu and Zhibo Chen are with the School of Information Science and Technology, University of Science and Technology of China, Hefei, 230026, China (e-mail: hanxinzhu@mail.ustc.edu.cn, chenzhibo@ustc.edu.cn).
}
\thanks{
Jiayi Luo is with Beihang University, Beijing 100191, China, and also with Zhongguancun Academy, Beijing 100094, China (e-mail: luojy@buaa.edu.cn).
}
\thanks{
Yonglin Tian, Ruiqi Song, Boyi Sun and Long Chen are with the State Key Laboratory of Multi-modal Artificial Intelligence Systems, Institute of Automation, Chinese Academy of Sciences, Beijing 100190, China (e-mail: yonglin.tian@ia.ac.cn; ruiqi.song@ia.ac.cn; sunboyi2024@ia.ac.cn; long.chen@ia.ac.cn).
}
}

\markboth{Journal of \LaTeX\ Class Files,~Vol.~14, No.~8, August~2021}%
{Shell \MakeLowercase{\textit{et al.}}: A Sample Article Using IEEEtran.cls for IEEE Journals}


\maketitle

\begin{abstract}
    Video generation models have recently attracted substantial attention for their ability to generate visually compelling videos, yet ensuring physically consistent and plausible dynamics still remains a fundamental challenge, driving a growing line of research on physical realism in video generation.
    To address this challenge, motivated by the fact that physical regularities are primarily encoded in motion patterns, we propose \underline{PhysFlow}, a novel two-stage framework for improving the physical plausibility of generated videos by decomposing video generation into motion-aware optical flow generation followed by motion-conditioned appearance synthesis.
    Specifically, PhysFlow consists of a physics-aware optical-flow video generator called PA-Flow and a flow-guided video generator called FlowRender. 
    During the first stage, PA-Flow employs a physics-aware attention module to model how motion attributes and material properties influence global motion and local deformation, respectively, and generates an optical flow video as an explicit representation of motion. 
    In the second stage, FlowRender leverages the decoupled motion representation as guidance to synthesize realistic textures and appearances, ultimately producing the final physically plausible video. 
    To further support model training with explicit physical supervision, we construct PhysVideo, a physics-based video dataset generated with a physics engine and 3D-GS rendering, containing 10K foreground objects and 50K realistic video sequences with annotations of motion and material properties. 
    Extensive experiments demonstrate that our proposed PhysFlow generates videos with superior physical plausibility while maintaining high visual fidelity compared with existing methods. 
    Project page: \url{https://physwm.github.io/PhysFlow/}.
\end{abstract}


\section{Introduction}
\label{sec:intro}

\begin{figure}[t]
    \centering
    \includegraphics[width=1\linewidth]{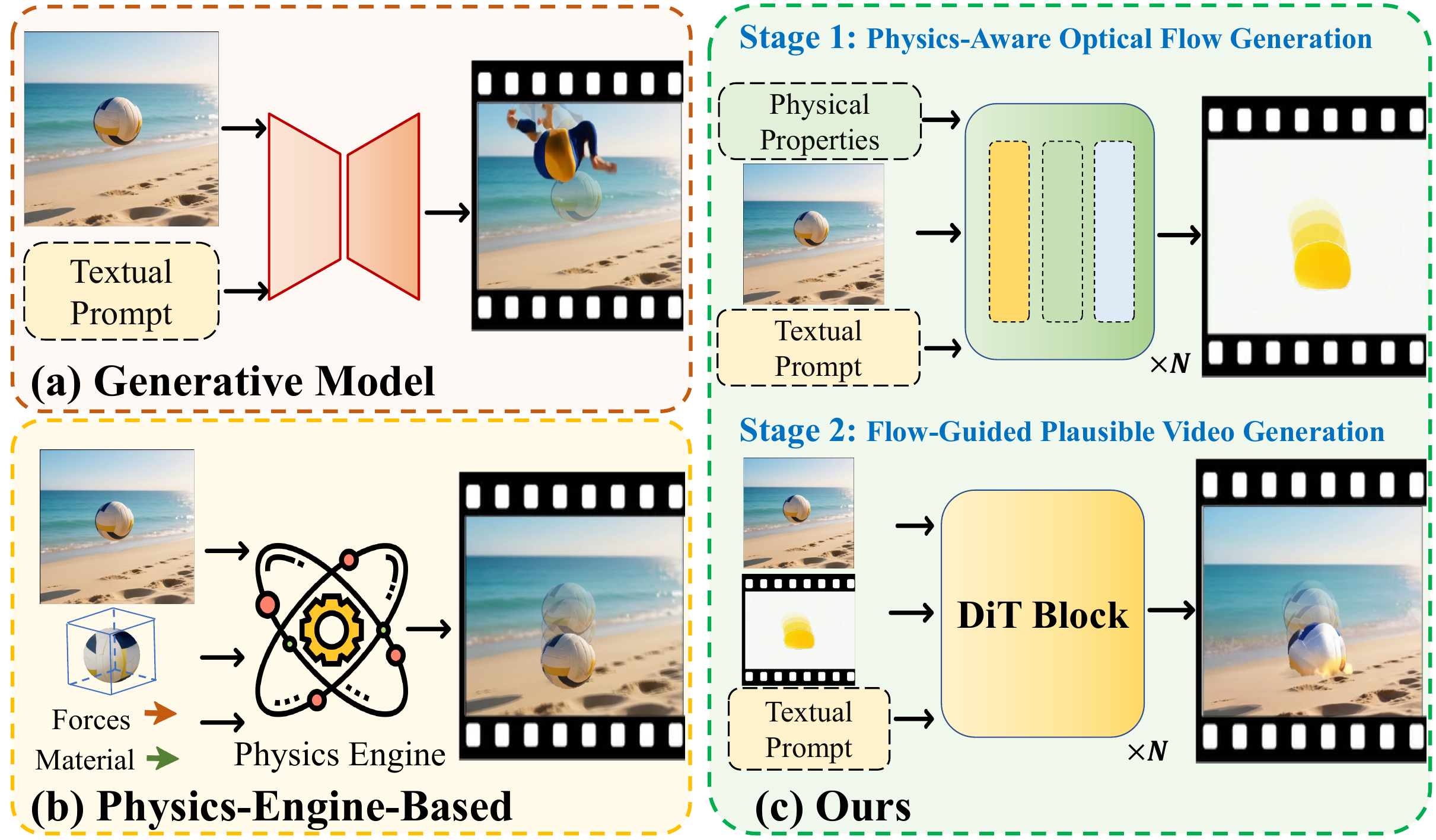}
    \caption{\textbf{Comparison of our method with prior video generation paradigms}. (a) Data-driven generative models. (b) Physics-engine-based methods. (c) Our physics-aware two-stage framework utilizing optical flow.}
    \label{fig:teaser}
\end{figure}

\IEEEPARstart{V}{ideo} generation models~\cite{blattmann2023stable, zheng2024open, kong2024hunyuanvideo, yangcogvideox, wan2025wan} have achieved remarkable progress in synthesizing high-quality and temporally coherent videos~\cite{nan2024openvid}. 
Despite their impressive visual quality, these models often lack a robust understanding of the underlying physical principles~\cite{bansal2024videophy, meng2024towards}. 
Consequently, the generated videos frequently exhibit physically implausible dynamics, such as inconsistent object deformation, unrealistic collisions, and incorrect interactions with the surrounding environment, which substantially limits their applicability to real-world scenarios that require accurate physical reasoning like embodied intelligence and autonomous driving~\cite{chi2025wow,mao2025robot,yang2026geniedrive}.

To equip video generation models with a stronger understanding of physical dynamics, recent efforts can be broadly categorized into two paradigms: Physics-engine-based approaches introduce explicit simulation into the generation pipeline, as illustrated in Figure~\ref{fig:teaser}. 
For example, PhysGen~\cite{liu2024physgen} uses rigid-body simulation to guide image-to-video generation, while PhysGen3D~\cite{chen2025physgen3d}, WonderPlay~\cite{li2025wonderplay}, OmniPhysGS~\cite{lin2025omniphysgs}, and related physics-engine-based methods~\cite{xie2024physgaussian, lin2025phys4dgen, liu2024physics3d} further combine 3D representations with physics solvers to support more complex dynamic scenes. These methods improve physical plausibility by relying on explicit geometry, material parameters, and solver-based state updates. However, their performance is often limited by the quality of 3D reconstruction, the accuracy of estimated physical parameters, and the fidelity of the simulator. As a result, errors in geometry, contact estimation, or material assignment may propagate to the final video, especially in scenes with non-rigid deformation, multi-object interaction, or complex foreground-background contact.

\textcolor{black}{A complementary paradigm seeks to endow pretrained video generators with physical controllability through additional conditioning signals. Force Prompting~\cite{gillman2025force} injects force signals into video generation, PhysCtrl~\cite{wang2025physctrl} represents dynamics with physics-conditioned 3D point trajectories, and recent physics-aware text-to-video systems~\cite{xue2025phyt2v, wang2025wisa} use language or world-simulation priors to improve physical realism. These approaches reduce direct dependence on full simulation at inference time, but they still face several limitations. Text-level controls are often too coarse to describe dense object deformation, while trajectory-based controls usually depend on foreground 3D structures and sparse motion points. Consequently, existing physics-aware methods still struggle to jointly achieve dense motion controllability, realistic appearance synthesis, and robust handling of object-background or multi-object interactions from a single image.}

Our key observation is that physical regularities are primarily manifested in motion patterns rather than textures~\cite{chi2025wow}. 
Motivated by this insight, we advocate learning physics-aware video generation by disentangling motion and appearance, instead of relying on physics engines that require manual settings.
We represent motion using optical flow, which compactly captures appearance-agnostic, pixel-level displacement~\cite{li2024generative}. 
\textcolor{black}{These limitations motivate us to use physics-aware optical flow as a dense 2D motion representation. Compared with simulator states or sparse 3D point trajectories, optical flow directly describes pixel-level motion in the image plane, providing a practical representation for local deformation, boundary-aware motion, and coupled object interactions while remaining compatible with modern video generators.}

Building on this idea, we propose \textbf{\underline{PhysFlow}}, a two-stage physics-aware video generation framework that decomposes video synthesis into (1) physics-aware optical flow generation and (2) flow-guided plausible video generation, as shown in Figure~\ref{fig:teaser} (c). 
The first stage is handled by the \textit{PA-Flow} module, a physics-aware optical-flow video generator that produces flow sequences conditioned on motion attributes (e.g., velocity and acceleration) and material properties (e.g., density and Young’s modulus). 
To learn how global motion and local deformations depend on these factors, PA-Flow employs a physics-aware attention module: motion attributes modulate global motion patterns, while material properties refine local deformation, enabling the model to learn complex non-rigid behaviors. 
The second stage is realized by the \textit{FlowRender} module, a flow-guided video generator that takes the optical-flow video as motion guidance and synthesizes the corresponding RGB sequence, enriching it with realistic textures.
\textcolor{black}{In this work, we focus on scenarios involving rigid bodies, elastic deformation, and selected flexible bodies, in which the underlying physics is primarily manifested through motion.}
To support the physics-aware learning process, we further construct \textit{PhysVideo}, a physics-based video dataset built using a physics engine and 3D-GS rendering. 
\textcolor{black}{Specifically, PhysVideo comprises 10K foreground objects spanning rigid, elastic, and flexible categories, together with 50K realistic video sequences, each annotated with motion attributes and material properties.}
Extensive experimental results demonstrate that our proposed PhysFlow generates videos with more physically plausible dynamics, outperforming existing methods.

In summary, our core contributions are as follows:
\begin{itemize}
    \item We propose a physics-aware dynamic video generation framework named \textbf{PhysFlow} that adopts a two-stage architecture to disentangle motion and texture synthesis. 
    \item We introduce \textit{PA-Flow}, a physics-aware image-to-flow generation model that incorporates a physics attention mechanism to model how motion attributes and material properties affect global motion and local deformation.
    \item We develop \textit{FlowRender}, a flow-guided video generation model that synthesizes dynamic RGB videos under the guidance of optical flow sequences. 
    \item We construct \textit{PhysVideo}, a physics-based video dataset that ensures physical accuracy and visual realism through physics simulation and 3D-GS rendering, containing over 10K objects and 50K video samples, providing comprehensive training data for physics-aware video generation.
\end{itemize}

\section{Related Works}
\label{sec:relat}
\subsection{Controllable Video Generation}

Video generation models trained on large-scale text–video paired datasets have demonstrated their remarkable capabilities in synthesizing high-quality videos~\cite{ho2022video, blattmann2023stable, kong2024hunyuanvideo, yangcogvideox}. 
Previous studies have shown that pretrained models can be additionally guided by various control signals, including camera motion~\cite{fu20243dtrajmaster, he2024cameractrl}, point trajectories~\cite{geng2025motion, gu2025diffusion, burgert2025go}, anchor-frame videos~\cite{chen2025stance, romero2025learning}, and unified or scenario-specific control adapters~\cite{wang2025uniadapter,wen2026panacea,yang2026expressive}, enabling more controllable video generation. 
These methods demonstrate that introducing structured control signals can substantially improve user control over generated content, but the controls are usually defined from visual or geometric cues rather than physical causes. 
Meanwhile, other research efforts have explored leveraging different modalities to guide motion-aware video synthesis. Some approaches~\cite{li2024generative, jin2025flovd} employ optical flow videos as guidance, using flow maps to characterize motion dynamics and subsequently synthesize RGB videos based on them. Other methods~\cite{liang2024flowvid, lv2024gpt4motion} use depth videos, transition frames, or inter-frame motion reuse~\cite{zhang2025tvg,wang2025denoisingreuse} as conditioning or temporal cues to guide the generation of coherent video sequences. In addition, several text-driven approaches~\cite{xue2025phyt2v} attempt to infer more fine-grained descriptions of motion dynamics from textual instructions, and recent evaluation efforts further emphasize motion-centered video quality assessment\textcolor{black}{~\cite{zhang2026mogenvd}}, thereby encouraging more precise control over motion behaviors during video generation. 
\textcolor{black}{However, although these approaches yield gains in temporal coherence and controllability, they typically treat motion as a visual trajectory to be followed rather than as the outcome of material properties and external motion conditions.}
Moreover, they generally lack explicit modeling of physical laws and often produce results that violate basic principles of physical plausibility.
To address these aforementioned limitations, our method generates physics-aware optical flow video as a motion representation and subsequently renders it into plausible video outputs.

\subsection{Physics-Engine-Based Video Generation}

Physics engines typically use the 3D representation of foreground objects as input to simulate and generate the 3D state at each time step, based on defined material and motion properties~\cite{xie2024physgaussian, chen2025physgen3d}. Leveraging the renderable 3D-GS representation~\cite{kerbl20233d}, some methods~\cite{xie2024physgaussian} propose using the Material Point Method (MPM) solver to update the 3D Gaussian representation at each time step after force-induced motion. However, due to the limitations of MPM, it is only suitable for simulating elastic objects. To address the shortcomings of MPM solvers, some approaches~\cite{feng2024gaussian, gao2025fluidnexus} incorporate Position-Based Dynamics (PBD) solvers to simulate fluid motion. The use of physics engines requires additional definition of physical properties, and some methods~\cite{zhao2025physsplat, mao2025live} propose using MLLMs to estimate properties such as Young's modulus, Poisson's ratio, and density. Furthermore, other methods~\cite{huang2025dreamphysics, lin2025omniphysgs, zhang2024physdreamer, zhu2026cp4d} suggest utilizing the prior knowledge of pre-trained video generation models to iteratively optimize the material properties of objects through SDS loss~\cite{poole2022dreamfusion} after rendering the video.
Recent advances in text-guided 3D-aware generation and Gaussian-based image-to-3D reconstruction~\cite{cheng2025efficient,zhang2025humanrefgs} have further improved the accessibility of 3D priors for such pipelines. However, these representations must still be converted into physically valid simulation states before reliable dynamics can be produced. While physics-engine-based methods can ensure physical plausibility, they require manual specification of simulation conditions and depend on high-quality 3D representations, which limits their ability to generate high-quality videos from a single image.

\subsection{Physics Enhanced Video Generation}

\begin{figure*}[t]
    \centering
    \includegraphics[width=1\linewidth]{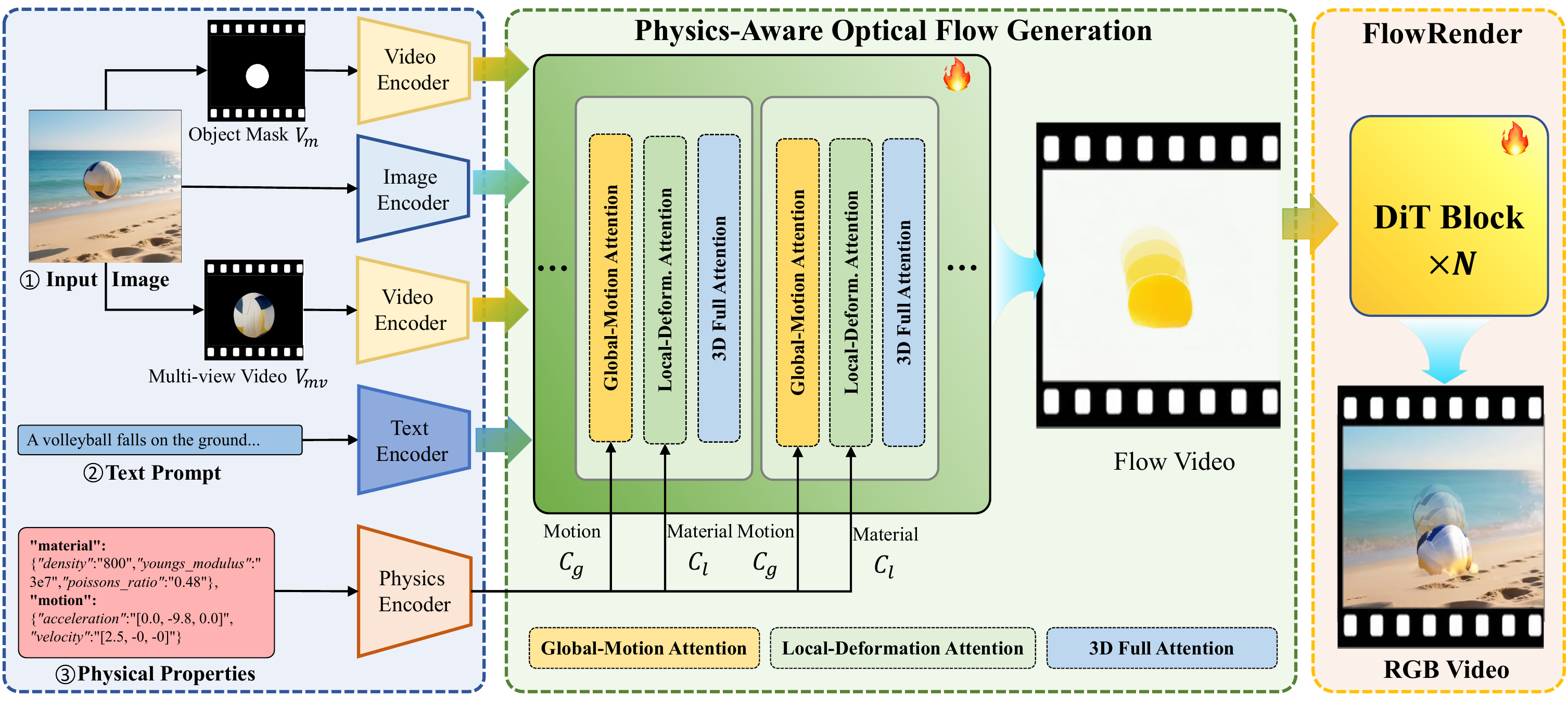}
    \caption{\textbf{Pipeline of our proposed PhysFlow}. PhysFlow generates physically plausible video via a two-stage pipeline: (1) generating a physics-aware optical flow video (Details are in Section~\ref{sec:pcflow}), (2) generating a plausible video with the guidance of the optical flow video (Details are in Section~\ref{sec:flowrender}).}
    \label{fig:pipeline}
\end{figure*}

Recent studies have increasingly focused on enhancing the physical plausibility of generated videos~\cite{wang2026physics}. WISA~\cite{wang2025wisa} introduces a dataset emphasizing physical phenomena, featuring complex interactions between fluids and solids. To assess physical realism, several works~\cite{bansal2024videophy, meng2024towards} propose VLM-based benchmarks and metrics for evaluating the physical quality of generated videos. To improve physical consistency, Force Prompting~\cite{gillman2025force} explicitly incorporates external forces and learns their influence on object motion. 
WonderPlay~\cite{li2025wonderplay} generates coarse videos with a physics engine and refines them via video optimization, but it still relies heavily on the simulator. In contrast, PhysCtrl~\cite{wang2025physctrl} reduces simulator dependence by learning a physically controllable point-cloud trajectory generator and using sparse motion trajectories to guide video synthesis. However, its performance is sensitive to the quality of pre-generated 3D point clouds, and it struggles with complex multi-object and object–background interactions.
Furthermore, NewtonGen~\cite{yuan2025newtongen} models physical dynamics by directly learning the governing partial differential equations, providing a principled formulation to guide video generation.
\textcolor{black}{While these methods simplify the cumbersome steps of physics engine simulations, they still rely on high-quality 3D object priors or overfitting to single scenes, making it challenging to achieve physics-aware video generation that accounts for both motion and material properties.}

\section{Methodology}
\label{sec:method}
\noindent\textbf{Task Description.} 
Given a single image, a textual prompt, and physical attributes as input, our goal is to generate videos with physically plausible object-centric dynamics that reflect the specified motion and material conditions.
We focus on rigid-body motion, elastic deformation, and selected flexible-body behaviors. Generating such videos requires capturing physics-conditioned object dynamics and synthesizing the corresponding motion-consistent changes in scene appearance.

\noindent\textbf{Method Overview.} 
To achieve the aforementioned goal, we propose a two-stage video generation framework that represents motion states through optical flow video, as shown in Figure~\ref{fig:pipeline}. We first train a physics-aware optical flow generator PA-Flow, which uses motion and material attributes as physical conditions to generate plausible global motion and local deformation (Section~\ref{sec:pcflow}). Subsequently, we train a flow-guided texture rendering model FlowRender to synthesize realistic appearances guided by the generated optical flow (Section~\ref{sec:flowrender}). To support the training of PA-Flow, we construct a physically realistic video dataset PhysVideo using a physics engine (Section~\ref{sec:dataset}). Finally, we introduce the training strategy of PA-Flow and FlowRender (Section~\ref{sec:train}).

\subsection{Stage \uppercase\expandafter{\romannumeral1}: Physics-Aware Optical Flow Generation}\label{sec:pcflow}

Given the input containing a static RGB image $I$, a textual prompt $P$, and a set of physical attributes comprising material parameters $M$ and motion parameters $(a, v)$, our goal is to synthesize an optical-flow video sequence that captures physics-grounded motion dynamics.

To enable more realistic motion generation, we introduce additional conditional priors to guide the video synthesis process. Specifically, we employ Grounded-SAM~\cite{ren2024grounded} to automatically segment foreground objects that correspond to the instructed motion, producing spatial masks that are temporally aligned to form a static mask video $\mathbf{V}_m$. To supplement object appearances that are not visible in the input viewpoint, we incorporate a multi-view object prior $\mathbf{V}_{mv}$ obtained from Trellis~\cite{xiang2025structured} as additional guidance. As illustrated in Figure~\ref{fig:pipeline}, the priors $\mathbf{V}_m$ and $\mathbf{V}_{mv}$ are jointly fed into PA-Flow as control signals to ensure semantic and appearance consistency. 
The input image $I$ is encoded into $\mathbf{H}_i$, while the conditional videos $\mathbf{V}_m$ and $\mathbf{V}_{mv}$ are encoded into $\mathbf{H}_m$ and $\mathbf{H}_{mv}$, respectively. After passing $\mathbf{V}_m$ and $\mathbf{V}_{mv}$ through the video projector, their processed latents are fused with $\mathbf{H}_i$ to construct the final conditioned latent $\mathbf{H}_c$.

\begin{figure}
    \centering
    \includegraphics[width=1\linewidth]{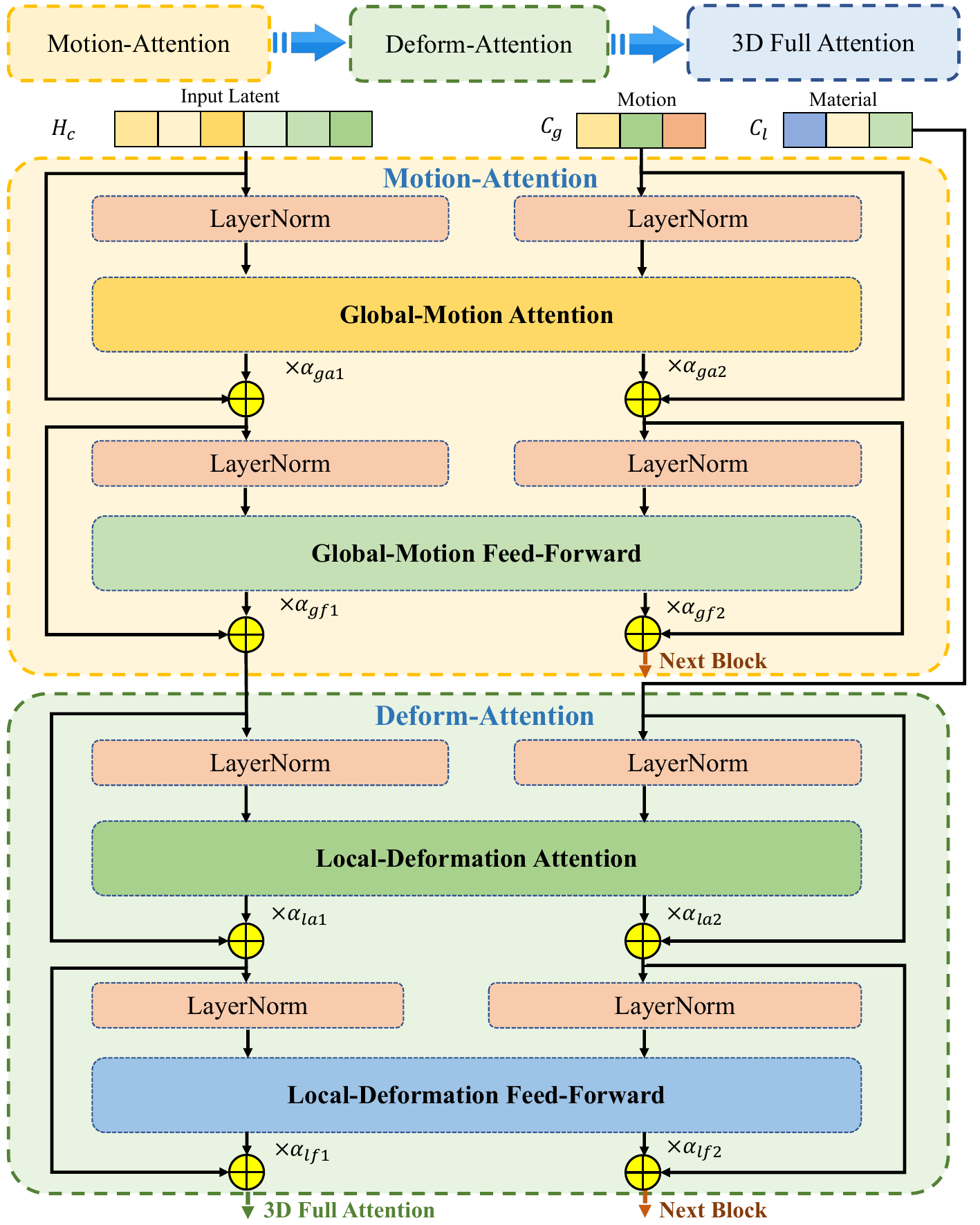}
    \caption{\textbf{The physics-aware attention module.} It consists of a global-motion attention module (Motion-Attention) and a local-deformation attention (Deform-Attention) within each block.}
    \label{fig:flow}
\end{figure}

The global motion of an object is primarily governed by motion-related attributes such as initial velocity and acceleration, whereas local deformation is further influenced by material properties, including Young’s modulus and density. To better enable physical attributes to guide the generation of optical flow videos, we design our model based on CogVideoX-I2V-5B~\cite{yangcogvideox} by introducing an additional physics-aware attention module (PAAM) into its original transformer blocks. As illustrated in Figure \ref{fig:flow}, PAAM comprises two key components: a global motion attention module driven by motion attributes, and a local deformation attention module driven by material attributes. The two modules are integrated in a sequential architecture, where the model first focuses on capturing global motion and subsequently learns local deformation patterns.

\noindent\textbf{Global-Motion Attention.}
The global motion attributes are defined by two components: the acceleration vector $\mathbf{a} = [a_x, a_y, a_z]$ and the velocity vector $\mathbf{v} = [v_x, v_y, v_z]$. 
During encoding, these vectors are extracted from the input motion dictionary and jointly embedded to form the global motion condition $\mathbf{C}_g$. The input latent $\mathbf{H}_c$ is fused with $\mathbf{C}_g$ through the global motion attention module to yield the motion-aware latent $\mathbf{H}_{ga}$ and an updated motion condition $\mathbf{C}_{ga}$. The process adapts residual connection and computes $\mathbf{H}_{ga}$ as:
\begin{equation}
    \mathbf{H}_{ga} = \mathbf{H}_c + \alpha_{ga1} \cdot \operatorname{Attn}_g(\operatorname{LN}([\mathbf{H}_c; \mathbf{C}_g])),
    \label{eq:h_ga}
\end{equation}
where $\alpha_{ga1}$ is a learnable scaling parameter, and $\mathrm{LN}$ denotes an adaptive layer normalization operation. The updated motion condition $\mathbf{C}_{ga}$ is computed in a similar manner. 
Finally, a feed-forward layer is applied, following the same residual design, to further refine the latent representation. The updated latent $\mathbf{H}_{gf}$ is computed as:
\begin{equation}
    \mathbf{H}_{gf} = \mathbf{H}_{ga} + \alpha_{gf1} \cdot \operatorname{FFN}_g(\operatorname{LN}([\mathbf{H}_{ga}; \mathbf{C}_g]))
\end{equation}
where $\alpha_{gf1}$ denotes a learnable scaling coefficient that controls the contribution of the feed-forward operation.

\noindent\textbf{Local-Deformation Attention.}
The material attributes are expressed as a vector $\mathbf{m}=[\rho, E, \nu]$, consisting of density $\rho$, Young’s modulus $E$, and Poisson’s ratio $\nu$.
This vector is extracted from the input material dictionary and encoded into a material condition embedding $\mathbf{C}_l$.
To model material-dependent deformation, $\mathbf{C}_l$ interacts with the motion-aware latent $\mathbf{H}_{gf}$ through the local deformation attention module, yielding the intermediate feature $\mathbf{H}_{la}$.
\begin{equation}
\mathbf{H}_{la} = \mathbf{H}_{gf} + \alpha_{la1} \cdot \operatorname{Attn}_l(\operatorname{LN}([\mathbf{H}_{gf}; \mathbf{C}_l])),
\end{equation}
where $\alpha_{la1}$ is a learnable coefficient controlling the strength of local deformation attention.
Subsequently, a feed-forward refinement is performed to further enhance material sensitivity, formulated as:
\begin{equation}
\mathbf{H}_{lf} = \mathbf{H}_{la} + \alpha_{lf1} \cdot \operatorname{FFN}_l(\operatorname{LN}([\mathbf{H}_{la}; \mathbf{C}_l])),
\end{equation}
where $\alpha_{lf1}$ denotes the learnable scaling factor in the feed-forward layer.

From the above process, a physics-aware latent is obtained and then processed by 3D full attention~\cite{yangcogvideox} and a feed-forward network to produce the predicted noise:
\begin{equation}
    \tilde{\mathbf{F}} = \operatorname{FFN}(\operatorname{Attn}([\mathbf{H}_{lf};\mathbf{C}_P])),
\end{equation}
where $\mathbf{C}_P$ denotes the encoded embedding of the textual prompt $P$.
The final predicted optical flow latent is calculated with the denoising processor $D_{\boldsymbol{\theta}}(\cdot)$:
\begin{equation}
    \hat{\mathbf{F}} = D_{\boldsymbol{\theta}}(\tilde{\mathbf{F}}_t, t),
\end{equation}
where $\tilde{\mathbf{F}}_t$ is the sampled noise at the timestamp $t$.

\subsection{Stage \uppercase\expandafter{\romannumeral2}: Flow-Guided Plausible Video Generation}\label{sec:flowrender}
In this stage, our objective is to generate a visually plausible video that adheres to the motion guidance provided by $\hat{F}$.
Since object motion inevitably affects the surrounding environment, this process involves more than merely moving the foreground while keeping the background static. 
However, explicitly modeling the complex interactions between the foreground and background remains challenging. To address this, we leverage the extensive prior knowledge embedded in large-scale pre-trained image-to-video generation models.
In particular, we introduce FlowRender, which is built upon CogVideoX-I2V-5B\cite{yangcogvideox} and trained via Supervised Fine-Tuning (SFT) under the conditioning of optical flow videos. The training process leverages real-world datasets such as OpenVid\cite{nan2024openvid}. To enhance the ability to generate complex physics-related phenomena, we further utilize the physics-focused dataset WISA~\cite{wang2025wisa} to fine-tune FlowRender.
The effects of fine-tuning on WISA~\cite{wang2025wisa} (Phys-FT) are illustrated in the ablation study (Table~\ref{tab:ablation1}).
FlowRender employs $\hat{\mathbf{F}}$ as the motion control condition:
\begin{equation}
    \tilde{\mathbf{R}} = \operatorname{FFN}(\operatorname{Attn}([\mathbf{H}_i; \hat{\mathbf{F}}; \mathbf{C}_P])),
\end{equation}
where $\tilde{\mathbf{R}}$ denotes the noise predicted by the model. 
The final output video latent $\hat{\mathbf{R}}$ is calculated after denoising:
\begin{equation}
    \hat{\mathbf{R}} = D_{\boldsymbol{\theta}}(\tilde{\mathbf{R}}_t, t),
\end{equation}
where $\tilde{\mathbf{R}}_t$ is the sampled noise at timestamp $t$.
The output of FlowRender represents the RGB video frames synthesized under the motion constraints imposed by $\hat{\mathbf{F}}$.
During training, the model learns to align the motion implied by the optical-flow condition with the visual content generation process, enforcing temporal coherence and ensuring that object dynamics remain physically consistent throughout the sequence.

\subsection{PhysVideo Dataset}\label{sec:dataset}
To facilitate the training of our physics-aware optical flow video generation model, we construct PhysVideo, a dataset comprising 50K physics-grounded video samples. 
Existing physics-engine–based video datasets primarily use simplistic or synthetic backgrounds. In contrast, PhysVideo incorporates realistic background scenes while maintaining physically grounded simulations.
Specifically, we generate 10K foreground objects represented by 3D-GS using Trellis~\cite{xiang2025structured}, and further reconstruct 10 diverse backgrounds with 3D-GS representations that support explorable viewpoints. Following the design of OmniPhysGS~\cite{lin2025omniphysgs}, we employ TaiChi~\cite{hu2019taichi} as the physics engine and adopt the Material Point Method (MPM) as the physics solver to compute object state updates given physical properties. Leveraging the advantages of 3D-GS, we directly compose the foreground object $G_f$ and background $G_b$, rendering full frames from specified viewpoints. 
For each sample, the textual prompt, motion attributes, and material parameters used to control the simulation are automatically recorded. During post-processing, we employ MemFlow~\cite{dong2024memflow} to estimate optical flow videos from the simulated sequences and use Grounded-SAM~\cite{ren2024grounded} to obtain semantic masks. Finally, Trellis~\cite{xiang2025structured} is again utilized to generate multi-view videos of the segmented foreground objects. 
Through the above pipeline, we construct a dataset comprising 10K objects and 50K videos. By incorporating realistic 3D backgrounds, our dataset produces highly realistic rendered videos, as shown in Figure~\ref{fig:supp_dataset2}.
\textcolor{black}{More dataset details can be found in Appendix Section A in the Supplementary Material.}

\begin{figure}[t]
    \centering
    \includegraphics[width=1\linewidth]{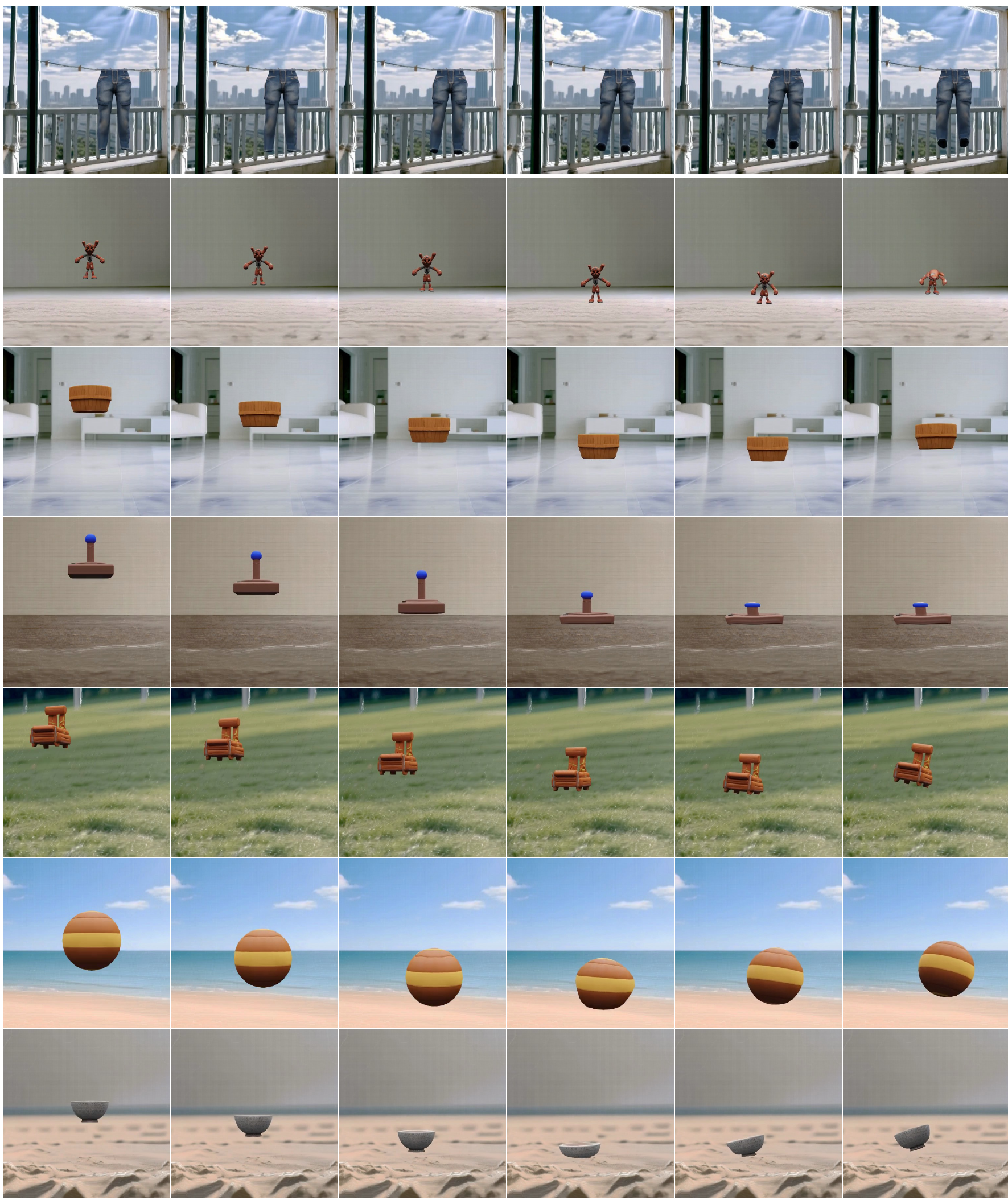}
    \caption{\textbf{Dataset demonstration of PhysVideo}. Our dataset contains various objects and motion patterns.}
    \label{fig:supp_dataset2}
\end{figure}

\subsection{Training Pipeline}\label{sec:train}

\noindent\textbf{PA-Flow Training.}
We optimize the model using the base SFT loss and incorporate physics-informed regularization to enforce spatial and temporal motion consistency. Specifically, the base loss is expressed as follows:
\begin{equation}
\label{eq:base-loss}
\mathcal{L}_{\mathrm{base}}
=
\mathbb{E}_{\tilde{\mathbf{F}},t}
\Bigl[w_1(t) \cdot
\bigl\|
D_{\boldsymbol{\theta}}(\tilde{\mathbf{F}}_t, t)
-
\bar{\mathbf{F}}
\bigr\|_2^2
\Bigr],
\end{equation}
where $\bar{\mathbf{F}}$ denotes the latent representation of the ground-truth optical-flow video, and $w_1(t)$ weights the denoising error at diffusion timestep $t$ according to the corresponding noise level.

To optimize the global motion, we introduce two complementary temporal regularization terms that constrain motion evolution in both the velocity and acceleration domains, namely the velocity-smoothness loss $\mathcal{L}_{\mathrm{vel}}$ and the acceleration-consistency loss $\mathcal{L}_{\mathrm{acc}}$. The velocity-smoothness loss penalizes abrupt temporal variations in the optical-flow sequence and is formulated as:
\begin{equation}
    \label{eq:vel-loss}
    \mathcal{L}_{\mathrm{vel}}
    = 
    \frac{1}{T-2}
    \sum_{t=2}^{T-1}
    \bigl\|
    \hat{\mathbf{F}}_{t+1}-2\hat{\mathbf{F}}_{t}+\hat{\mathbf{F}}_{t-1}
    \bigr\|_2^2,
\end{equation}
while the acceleration-consistency loss enforces alignment between the predicted and ground-truth temporal derivatives of optical flow:
\begin{equation}
    \label{eq:acc-consistency}
    \mathcal{L}_{\mathrm{acc}}
    =
    \frac{1}{T-1}
    \sum_{t=1}^{T-1}
    \bigl\|
    (\hat{\mathbf{F}}_{t+1}-\hat{\mathbf{F}}_t)
    -
    (\bar{\mathbf{F}}_{t+1}-\bar{\mathbf{F}}_t)
    \bigr\|_{1}.
\end{equation}

To optimize local deformation under physical constraints, we introduce an elastic strain energy loss derived from continuum mechanics, which models the internal potential energy stored in a deformable material under external stress:
\begin{equation}
    \label{eq:elastic-loss}
    \mathcal{L}_{\mathrm{ela}}
    \!=\!
    \frac{1}{T}\!
    \sum_{t=1}^{T}\!
    \Big[\!
    \frac{\mu}{2}\!\,\bigl\|\nabla\hat{\mathbf{F}}_{t}\!+\!
    \nabla\hat{\mathbf{F}}_{t}^{\top}\bigr\|_\mathrm{F}^{2}\!+ \!
    \frac{\lambda}{2}\!\,\bigl[\operatorname{tr}\bigl(\tfrac{\nabla\hat{\mathbf{F}}_{t}+
    \nabla\hat{\mathbf{\mathbf{F}}}_{t}^{\top}}{2})\bigr]^{2}
    \Big],
\end{equation}
where $\nabla\hat{\mathbf{F}}_{t}$ is the spatial Jacobian of the optical flow,  
$(\cdot)^{\top}$ denotes matrix transpose,  
$\mathrm{tr}(\cdot)$ is the trace operator,  
and $\|\cdot\|_\mathrm{F}$ is the Frobenius norm.  
The coefficients $\lambda$ and $\mu$ are Lamé parameters computed from the Young’s modulus $E$ and Poisson’s ratio $\nu$ as  
$\mu = \tfrac{E}{2(1+\nu)}$ and $\lambda = \tfrac{E\nu}{(1+\nu)(1-2\nu)}$.  
This term penalizes non-physical local deformations in the flow field, corresponding to minimizing the elastic strain energy.
Finally, the total training objective is formulated as follows:
\begin{equation}
\label{eq:total-loss}
\mathcal{L}_1
=
\mathcal{L}_{\mathrm{base}}
+ \lambda_{\mathrm{vel}}\mathcal{L}_{\mathrm{vel}}
+ \lambda_{\mathrm{acc}}\mathcal{L}_{\mathrm{acc}}
+ \lambda_{\mathrm{ela}}\mathcal{L}_{\mathrm{ela}},
\end{equation}
where $\lambda_{\mathrm{vel}}$, $\lambda_{\mathrm{acc}}$ and $\lambda_{\mathrm{ela}}$ are balancing coefficients that control the relative contribution of each term.

\begin{table*}[t]
    \centering
    \caption{\textbf{Quantitative comparisons.} \textbf{Bold}: Best. \underline{Underline}: Second Best. PE refers to the physics engine.}
    \setlength{\tabcolsep}{0.5pt}
    \renewcommand{\arraystretch}{1.0}
    \begin{tabular*}{\textwidth}{@{\extracolsep{\fill}} l c ccc cccc ccc @{}}
    \toprule
    \multirow{2}{*}{\textbf{Method}} & \multirow{2}{*}{\textbf{PE}} & \multicolumn{3}{c}{\textbf{VideoPhy-2}} & \multicolumn{4}{c}{\textbf{VBench}} & \multicolumn{3}{c}{\textbf{WorldScore}} \\
    \cmidrule(lr){3-5} \cmidrule(lr){6-9} \cmidrule(lr){10-12}
    &  & SA$\uparrow$ & PC$\uparrow$ & Rule$\uparrow$ & Motion$\uparrow$ & Subject$\uparrow$ & Flicker$\uparrow$ & Image$\uparrow$ & Photo$\uparrow$ & Motion$\uparrow$ & 3D$\uparrow$  \\ \midrule
        OmniPhysGS~\cite{lin2025omniphysgs} & \Checkmark & 2.41 & 3.09 & 0.204 & 0.995 & 0.919 & 0.993 & 0.410 & 12.41 & 89.47 & 39.85 \\
        PhysGen~\cite{liu2024physgen} & \Checkmark & 2.58 & 3.25 & 0.265 & 0.995 & \underline{0.958} & 0.990 & 0.629 & 89.60 & 80.65 & \underline{91.52} \\
        PhysGen3D~\cite{chen2025physgen3d} & \Checkmark & 2.61 & \underline{3.27} & 0.194 & \underline{0.996} & 0.928 & \textbf{0.998} & 0.593 & \underline{92.60} & \underline{90.07} & \textbf{92.12} \\
        CogVideoX-I2V-5B~\cite{yangcogvideox} & \XSolid & 2.66 & 3.14 & 0.221 & 0.992 & 0.917 & 0.988 & 0.628 & 73.22 & 74.48 & 80.49 \\
        Wan2.2-TI2V-5B~\cite{wan2025wan}& \XSolid & \underline{2.71} & 3.18 & 0.188 & 0.994 & 0.916 & 0.989 & \underline{0.635} & 59.86 & 60.25 & 74.06 \\
        Force Prompting~\cite{gillman2025force} & \XSolid & 2.69 & 3.21 & 0.219 & 0.994 & 0.951 & 0.992 & 0.623 & 65.83 & 52.64 & 76.86 \\
        PhysCtrl~\cite{wang2025physctrl} & \XSolid & 2.68 & 3.22 & \underline{0.250} & 0.996 & 0.955 & 0.994 & 0.657 & 90.80 & 82.45 & 90.27 \\
        \midrule
        \textbf{PhysFlow(Ours)} & \XSolid & \textbf{2.81} & \textbf{3.41} & \textbf{0.281} & \textbf{0.997} & \textbf{0.960} & \underline{0.994} & \textbf{0.640} & \textbf{95.59} & \textbf{92.82} & 90.49 \\
        \bottomrule
    \end{tabular*}
    \label{tab:com}
\end{table*}

\noindent\textbf{FlowRender Training.}
To train FlowRender, we adopt a diffusion-based image-to-video learning paradigm in which the model learns to reconstruct video frames conditioned on the optical-flow guidance $\hat{\mathbf{F}}$.
Specifically, we minimize the mean squared error between the denoised prediction and the ground-truth frame under random diffusion steps:
\begin{equation}
    \label{eq:flowrender-loss}
    \mathcal{L}_{2}
    = 
    \mathbb{E}_{\tilde{{\mathbf{R}}},\,t}
    \Bigl[w_2(t)
    \bigl\|
    D_{\boldsymbol{\theta}}(\tilde{\mathbf{R}}_t, t)
    -
    \bar{\mathbf{R}}
    \bigr\|_2^2
    \Bigr],
\end{equation}
where $\bar{\mathbf{R}}$ is the ground-truth RGB video.
This training objective guides FlowRender to synthesize temporally coherent and visually realistic video sequences that faithfully follow the flow-based motion guidance.

\section{Experiments}
\label{sec:exp}

\begin{figure*}[t]
    \centering
    \includegraphics[width=1\linewidth]{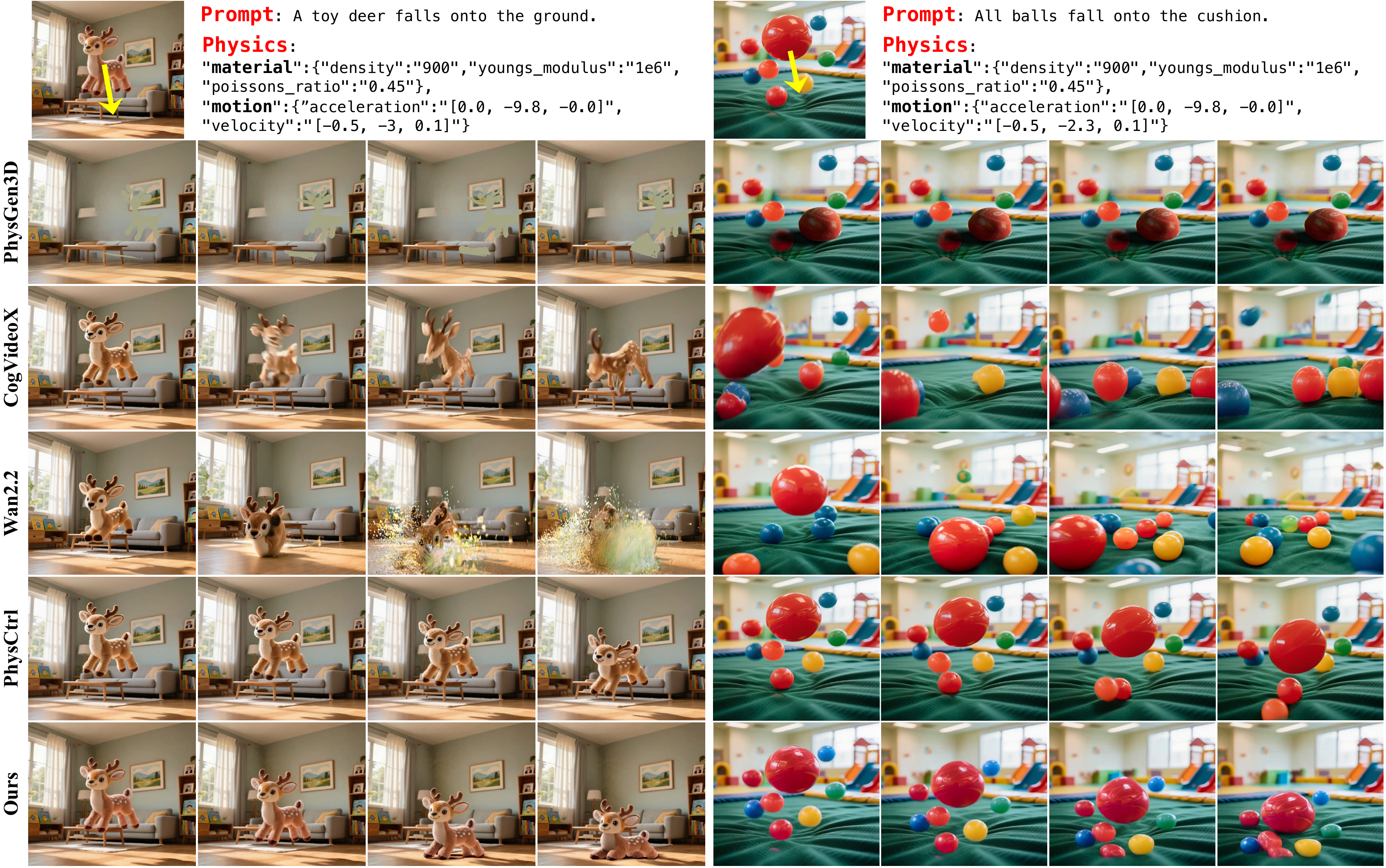}
    \vspace{-0.4cm}
    \caption{\textbf{Qualitative comparisons.} The top row shows the given image, text prompt and corresponding physics settings. For the definition of velocity and acceleration directions, the positive x-axis is defined as horizontal left, the positive y-axis as vertical upward, and the positive z-axis as perpendicular to the paper plane inward.}
    \label{fig:com}
    \vspace{-0.3cm}
\end{figure*}

\subsection{Experimental Setups}
\noindent\textbf{Implementation details.}
We train our model on 8 NVIDIA A100 GPUs with 80GB GPU memory. All inference experiments are performed on a single A100 GPU.
To verify the performance of our proposed method, we construct an evaluation dataset comprising \textcolor{black}{60} samples, each containing an image $I$ with a resolution of $512\times512$, a corresponding textual prompt $P$, and physical properties provided by GPT-4o.
\textcolor{black}{The evaluation dataset includes rigid bodies, elastic objects, and selected flexible objects, together with scenarios involving complex-background contact and multi-object interaction.}
The image $I$ is synthetically generated in a realistic style using a text-to-image generation model~\cite{wu2025qwen}, and the evaluation also includes real-world natural images.
The foreground mask $\mathbf{M}_f$ is derived from $I$ via Grounded-SAM\cite{ren2024grounded}. Subsequently, the foreground multi-view video and corresponding 3D-GS representation are synthesized with Trellis~\cite{xiang2025structured}, facilitating the simulation of physical-engine based baseline methods.

\noindent\textbf{Baselines.}
\textcolor{black}{We compare our method with representative baselines from two complementary families: physics-engine-based methods and video generative models. The first family evaluates whether explicit simulation pipelines can produce physically plausible dynamics when provided with reconstructed foreground geometry and scene priors. We therefore include PhysGen~\cite{liu2024physgen}, PhysGen3D~\cite{chen2025physgen3d}, and OmniPhysGS~\cite{lin2025omniphysgs}, which rely on physics simulation and 3D scene representations to drive object motion. The second family examines the behavior of data-driven video generators under the same image and prompt conditions. We evaluate general-purpose models, including CogVideoX~\cite{yangcogvideox} and Wan~\cite{wan2025wan}, to measure the physical consistency achievable by large-scale generative priors alone. In addition, we include Force Prompting~\cite{gillman2025force} and PhysCtrl~\cite{wang2025physctrl} as physics-enhanced generative baselines, since they introduce explicit force or trajectory controls to improve dynamic plausibility. This selection covers both simulator-driven and learned generation paradigms, enabling a comprehensive comparison across visual fidelity, temporal coherence, and physics-aware controllability.}

\noindent\textbf{Metrics.}
We evaluate generated videos using complementary objective metrics and human evaluation. VideoPhy-2~\cite{bansal2025videophy2} reports semantic adherence (SA), which measures agreement with the input prompt; physical commonsense (PC), which assesses holistic compliance with real-world physics; and physical-rule adherence (Rule), which measures whether prompt-specific physical rules are satisfied. WorldScore~\cite{duan2025worldscore} evaluates photo consistency (Photo), 3D consistency (3D), and motion smoothness (Motion), while VBench~\cite{huang2024vbench} measures motion smoothness (Motion), subject consistency (Subject), temporal flickering (Flicker), and image quality (Image). 
We additionally report the Mechanics dimension from VBench-2.0~\cite{zheng2025vbench2}, which evaluates whether generated dynamics follow basic mechanical principles such as gravity and stress.
Following prior works~\cite{chen2025physgen3d, li2025wonderplay}, we further evaluate the generated videos with GPT-4o along three dimensions: physical realism (Physical), photorealism (Photo), and semantic consistency (Semantic). To complement these model-based evaluations, we report Phys., a human-rated physical-plausibility score aggregated from 50 assessments. Higher values indicate better performance across all metrics.

\subsection{Comparisons with State-Of-The-Art Methods}

\noindent\textbf{Quantitative Evaluation}. 
Quantitatively, as shown in Table~\ref{tab:com}, our method achieves superior motion coherence and temporal smoothness, consistently outperforming prior video generative models and shows comparable results with physics-engine-based methods across key dynamic metrics.
Moreover, the generated videos exhibit higher visual quality than physics-engine-based methods. Regarding physical plausibility, as demonstrated in Table~\ref{tab:gpt}, our method surpasses all competing approaches on the physics realism metric, while simultaneously maintaining strong alignment with the input text, thereby ensuring high semantic consistency.

\begin{table}
    \centering
    \caption{\textbf{Complementary Evaluation Results}. Physical, Photo, and Semantic are evaluated by GPT-4o; Mechanics is from VBench-2.0; and Phys. is based on 50 human assessments.}
    \setlength{\tabcolsep}{2.0pt}
    \renewcommand{\arraystretch}{1.0}
    \begin{tabularx}{\linewidth}{@{} l|ccc|c|c @{}}
    \toprule
        \multirow{2}{*}{\textbf{Method}} & \multicolumn{3}{c|}{\textcolor{black}{\textbf{GPT-4o}}} & \multicolumn{1}{c|}{\textcolor{black}{\textbf{VBench-2.0}}} & \textcolor{black}{\textbf{Human}} \\
        \cmidrule(lr){2-4} \cmidrule(lr){5-5} \cmidrule(lr){6-6}
        & Physical$\uparrow$ & Photo$\uparrow$ & Semantic$\uparrow$ & Mechanics$\uparrow$ & Phys.$\uparrow$\\
        \midrule
        OmniPhysGS~\cite{lin2025omniphysgs} & 0.467 & 0.293 & 0.433 & 0.316 & 0.564\\
        PhysGen~\cite{liu2024physgen} & 0.530 & 0.890 & 0.598 & \underline{0.417} & \underline{0.608} \\
        PhysGen3D~\cite{chen2025physgen3d} & \underline{0.597} & 0.768 & 0.594 & 0.387 & 0.552 \\
        CogVideoX-I2V-5B~\cite{yangcogvideox} &  0.570 &  0.875 & 0.709 & 0.403 & 0.566 \\
        Wan2.2-TI2V-5B~\cite{wan2025wan} & 0.567 & \underline{0.891} & \underline{0.733} & 0.362 & 0.560 \\
        Force Prompting~\cite{gillman2025force} & 0.525 & 0.886 & 0.559 & 0.354 & 0.544 \\
        PhysCtrl~\cite{wang2025physctrl} & 0.582 & 0.855 & 0.703 & 0.362 & 0.586 \\
        \midrule
        \textbf{PhysFlow(Ours)}  & \textbf{0.613} & \textbf{0.903} & \textbf{0.744} & \textbf{0.531} & \textbf{0.708} \\
        \bottomrule
    \end{tabularx}%
    \label{tab:gpt}
\end{table}

\begin{figure}[t]
    \centering
    \includegraphics[width=1\linewidth]{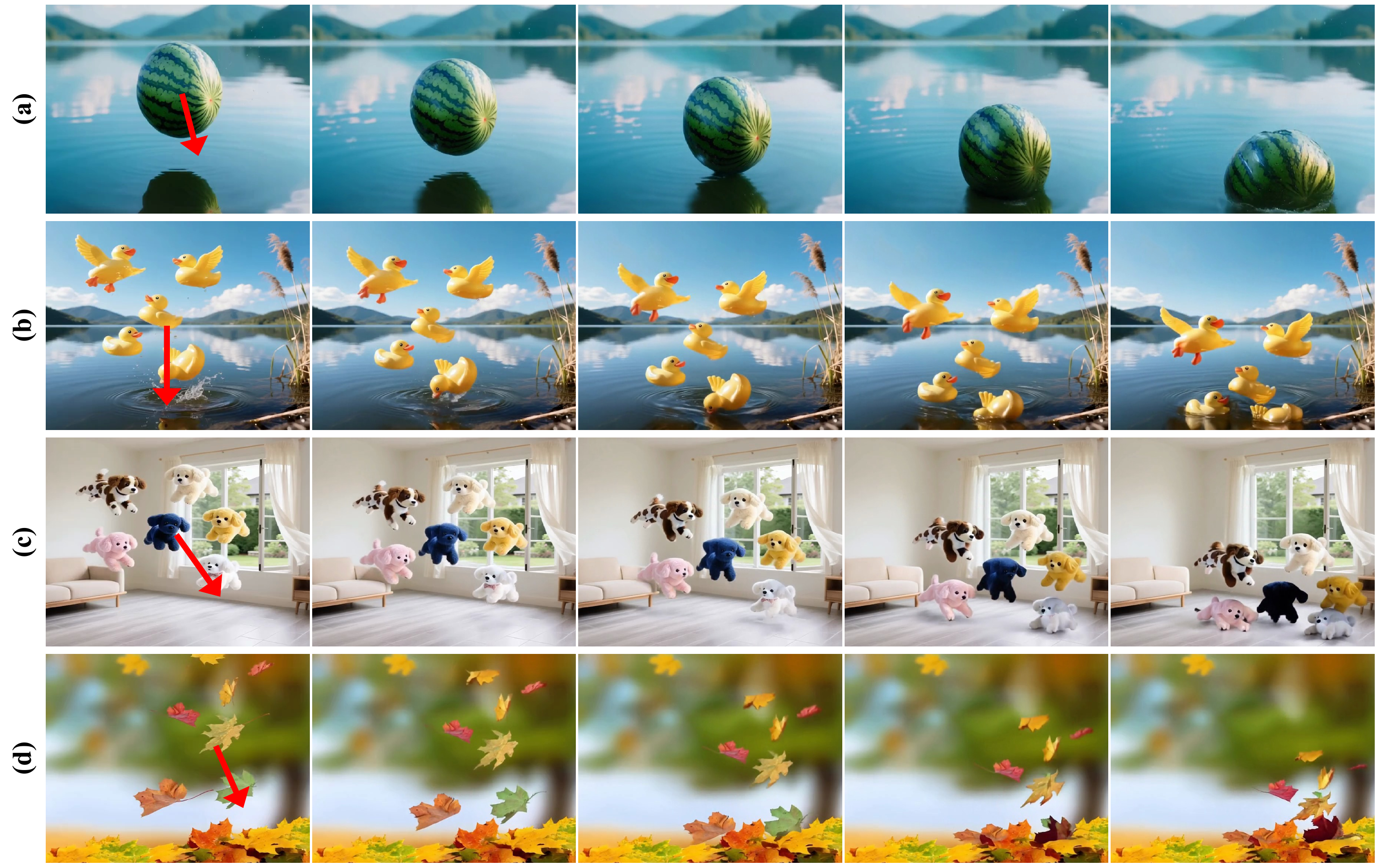}
    \caption{\textbf{Qualitative examples of challenging scenarios}. Each row shows a temporal sequence (left to right) under an applied external force $\mathbf{F}$ (red arrow), highlighting interactions with complex backgrounds and multi-object interactions.}
    \label{fig:show}
\end{figure}

\begin{figure}[t]
    \centering
    \includegraphics[width=1\linewidth]{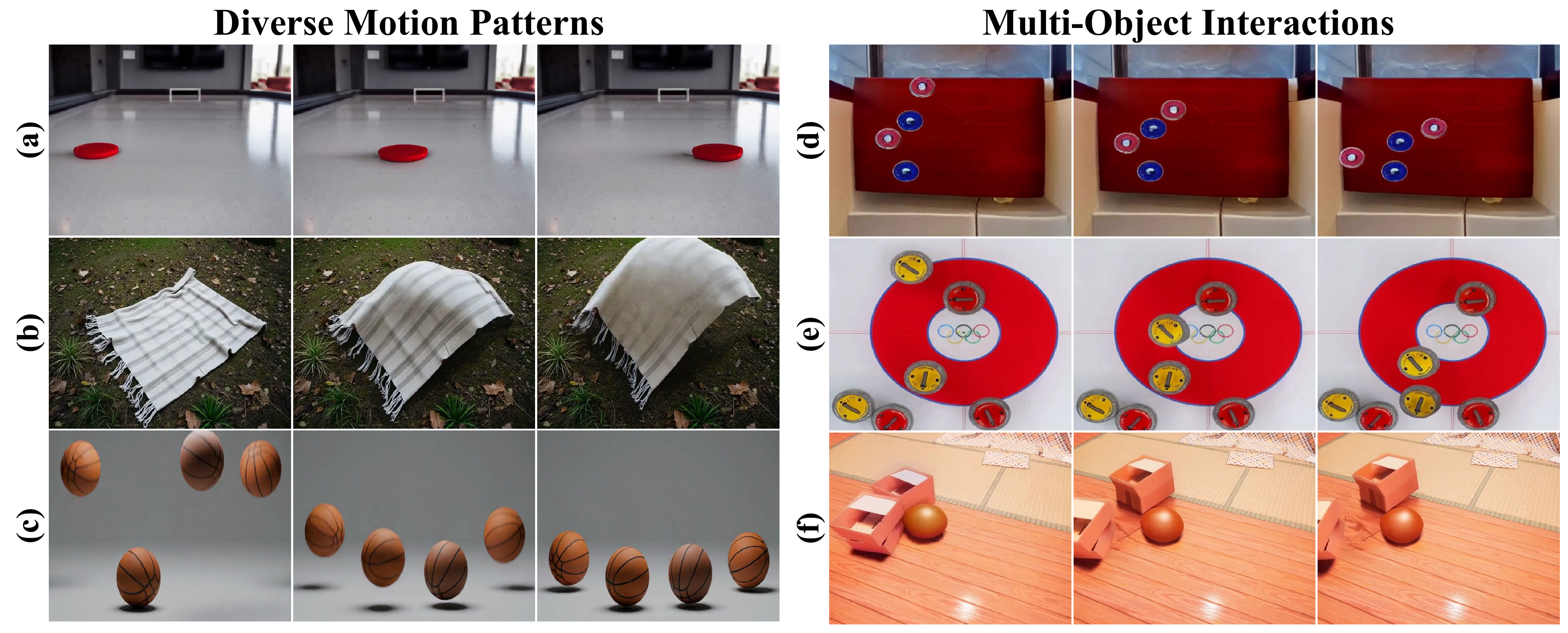}
    \caption{\textbf{Diverse motions and multi-object interactions.} PhysFlow generates diverse motion types and natural multi-object interactions.}
    \label{fig1}
\end{figure}

\begin{figure}[t]
    \centering
    \includegraphics[width=1\linewidth]{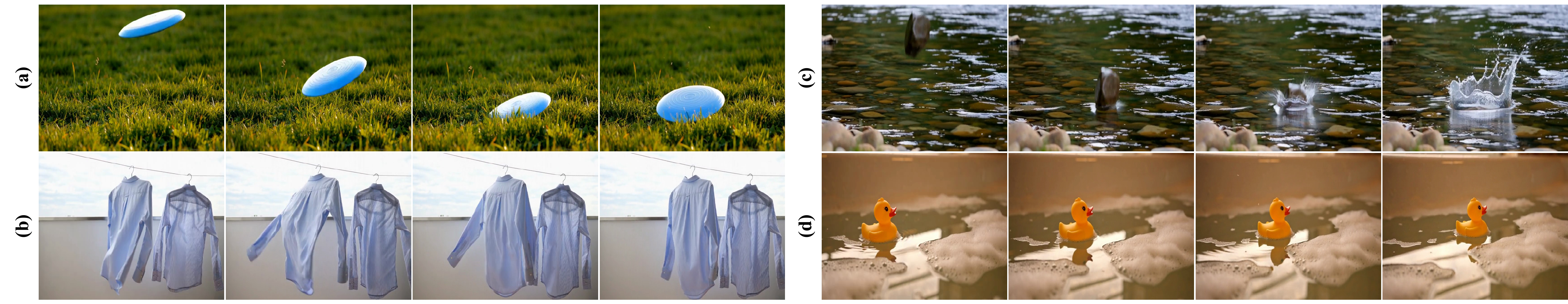}
    \vspace{-0.4cm}
    \caption{\textbf{Real-world generalization examples.} PhysFlow can generate natural motion in real-world environments.}
    \vspace{-0.3cm}
    \label{fig3}
\end{figure}

\noindent\textbf{Qualitative Evaluation}.
As illustrated in Figure~\ref{fig:com}, we present two cases for qualitative evaluation.
For the first scenario involving a non-rigid object with complex shape, PhysGen3D~\cite{chen2025physgen3d} suffers from texture loss on the foreground object due to insufficient mesh-based rendering. CogVideoX~\cite{yangcogvideox} fails to capture the correct motion dynamics, resulting in noticeable flickering. Wan~\cite{wan2025wan} can partially model the falling motion but misinterprets the material and interaction properties, causing the object to disperse like sand during descent. 
PhysCtrl~\cite{wang2025physctrl} misestimates the ground location, causing the foreground object to deform prematurely while still suspended in the air.
In contrast, our model accurately captures the falling behavior and produces physically plausible elastic deformations.
In the second scenario involving multi-object interactions, PhysGen3D~\cite{chen2025physgen3d} recognizes only one moving ball, leaving the others suspended in midair. CogVideoX~\cite{yangcogvideox} generates inconsistent motion after the fall, introducing several spurious red balls. 
Wan~\cite{wan2025wan} and PhysCtrl~\cite{wang2025physctrl} both fail to faithfully capture multi-object dynamics, with several balls remaining suspended in midair.
In comparison, our model successfully understands the multi-ball interaction dynamics, generating coherent motion for all objects.
To better demonstrate our method's performance on challenging scenarios, we present qualitative examples in Figure~\ref{fig:show}, covering interactions with fluid backgrounds as well as complex interactions among multiple objects.
More visual comparisons with baselines are provided in the \textcolor{black}{Appendix Section C in the Supplementary Material}.

\begin{figure*}[t]
    \centering
    \includegraphics[width=1\textwidth]{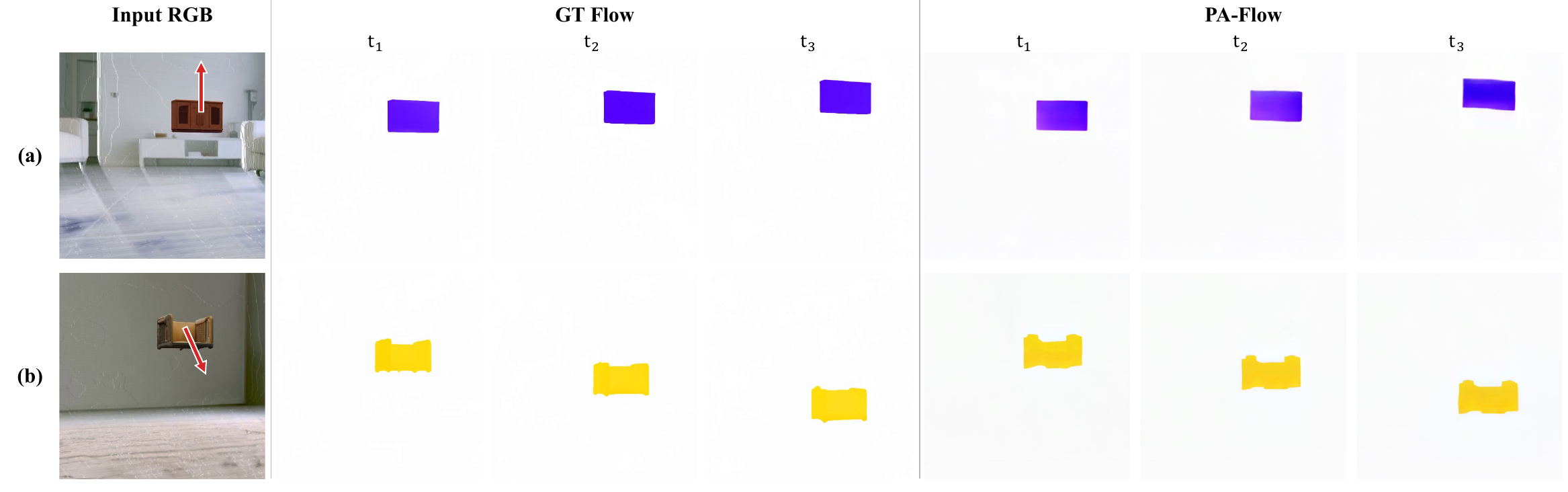}
    \caption{\textbf{Qualitative comparison of optical flow predictions}. The left column shows the input RGB image, while the middle and right panels show the ground-truth optical flow and PA-Flow predictions, respectively, at three matched time steps. PA-Flow produces motion patterns that closely follow the corresponding ground-truth flow.}
    \label{fig:stage1_flow_fidelity}
\end{figure*}

\begin{figure*}[t]
    \centering
    \includegraphics[width=1\textwidth]{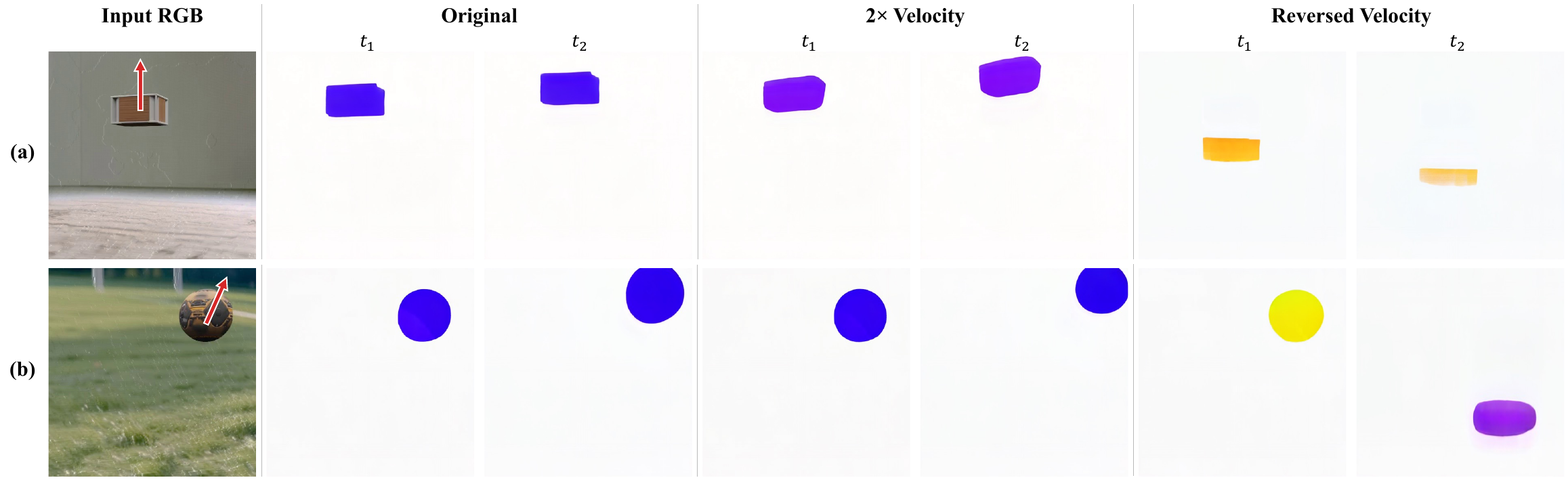}
    \caption{\textbf{Effects of initial velocity on PA-Flow predictions.} (a) and (b) show two representative examples. For each example, optical-flow predictions at two time steps are compared under the original, doubled-magnitude, and reversed-direction velocity settings, with all other conditions held fixed.}
    \label{fig:stage1_flow_control}
\end{figure*}

\begin{figure*}[t]
    \centering
    \includegraphics[width=1\linewidth]{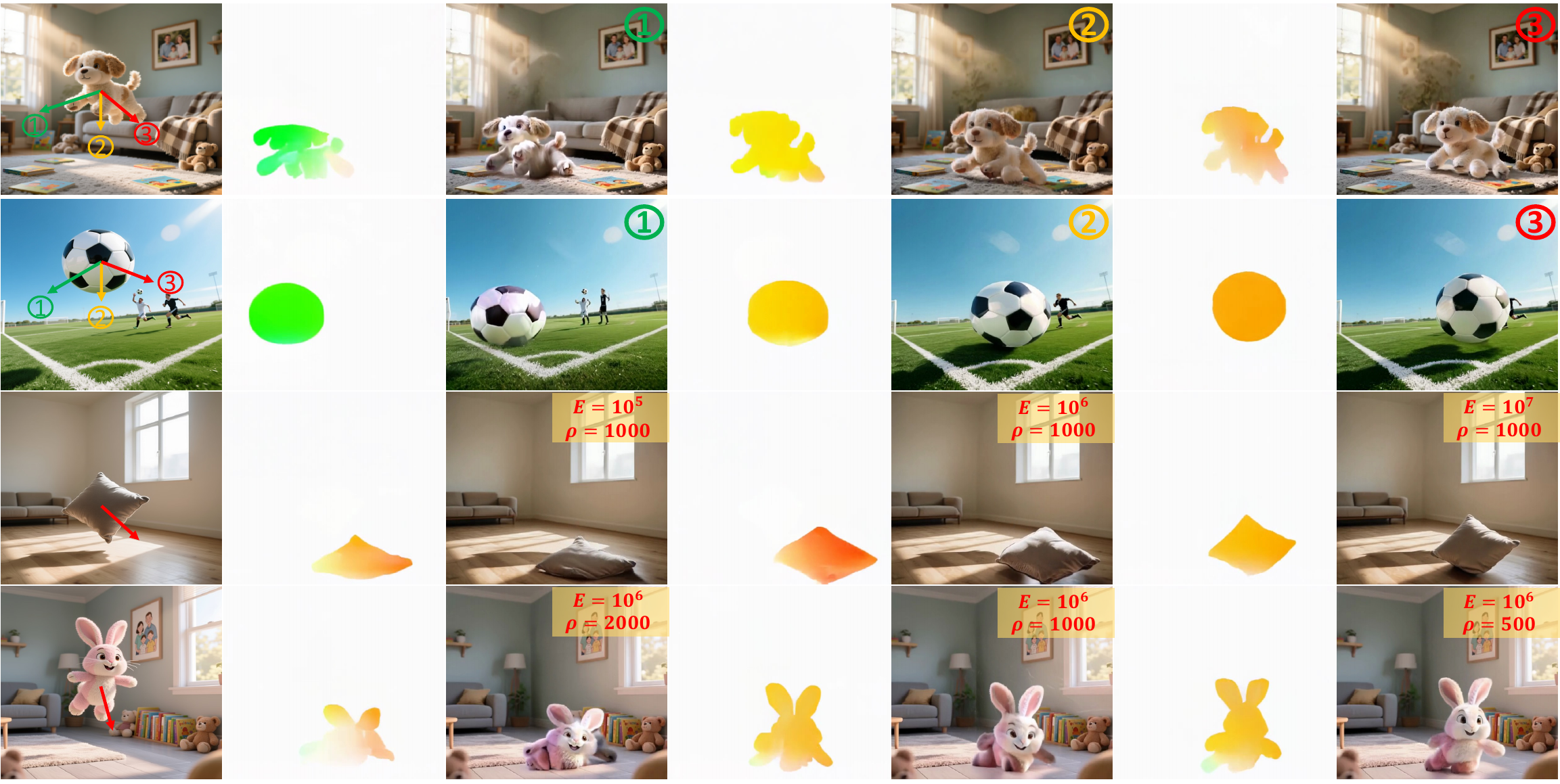}
    \caption{\textbf{Physics-conditioned flow-to-video generation.} The top two rows compare three initial-velocity directions indicated by the numbered arrows, whereas the bottom two rows vary Young's modulus and density, respectively. For each setting, the PA-Flow prediction is paired with a representative RGB frame from the corresponding FlowRender output, illustrating how motion and material conditions are reflected in the final video dynamics.}
    \label{fig:phys_control}
\end{figure*}

\noindent\textbf{Results on Varying Physical Properties.}
Real-world dynamic phenomena are fundamentally governed by physical laws, where variations in physical properties lead to distinct motion behaviors. 
In this analysis, we examine representative motion and material attributes, focusing on initial velocity, Young's modulus, and density.
To assess physical consistency, we visualize both the generated optical flow and the rendered videos under different physical configurations. Since our framework predicts optical flow before RGB rendering, we first inspect the intermediate PA-Flow output. 
Figure~\ref{fig:stage1_flow_fidelity} compares the ground-truth flow and PA-Flow predictions at matched time steps. Figure~\ref{fig:stage1_flow_control} then examines how PA-Flow responds to changes in initial velocity by holding all other conditions fixed and comparing the predicted optical flows under the original, doubled-magnitude, and reversed-direction velocity settings. 
Having established velocity-conditioned responses at the flow level, we next examine how physics-conditioned motion representations are reflected in the final visual output. Figure~\ref{fig:phys_control} jointly presents PA-Flow predictions and representative RGB frames from the corresponding FlowRender outputs. The top two rows vary the initial-velocity direction, whereas the bottom two rows vary Young's modulus and density, respectively. The paired results show that changes in motion and material conditions are reflected in both the intermediate optical flow and the rendered RGB output. In particular, changes in initial-velocity direction produce corresponding motion outcomes, while higher Young's modulus and lower density lead to smaller deformations, consistent with expected physical behavior.
\textcolor{black}{Beyond qualitative visualization, we further construct a physics-conditioned validation set to quantitatively evaluate controllability under varying physical properties. In this validation set, different methods are tested under controlled changes of motion and material attributes, allowing us to measure whether the generated videos remain visually coherent while following the specified physical conditions. As reported in Table~\ref{tab:quant_control}, our method achieves stronger physics-conditioned generation performance, further verifying its controllable behavior across different physical settings.}

\noindent{\textbf{Diverse motion and multi-object interaction results}.}
Figure~\ref{fig1} (a)--(c) shows that our method supports different motion types, including rigid sliding, cloth deformation, and bouncing. The observed variations indicate that the model captures object- and condition-dependent motion characteristics rather than producing a uniform motion response across scenarios. Figure~\ref{fig1} (d)--(f) show results on scenes with multiple interacting objects. Our method maintains temporally coherent motion across objects and produces plausible responses upon contact.

\begin{table}[t]
    \centering
    \caption{\textbf{\textcolor{black}{Quantitative comparison on physics-conditioned video generation.}}}
    \label{tab:quant_control}
    \setlength{\tabcolsep}{1.4pt}
    \renewcommand{\arraystretch}{1.0}
    \begin{tabular*}{\linewidth}{@{\extracolsep{\fill}} l|cccccc @{}}
        \toprule
        \textbf{Method} & PSNR$\uparrow$ & SSIM$\uparrow$ & LPIPS$\downarrow$ & FID$\downarrow$ & FVD$\downarrow$ \\
        \midrule
        PhysGen~\cite{liu2024physgen}  & 19.82 & 0.767 & 0.438 & 122.03 & 119.64 \\
        Force Prompting~\cite{gillman2025force}   & 21.08 & 0.847 & 0.316 & 105.22 & 103.47 \\
        PhysCtrl~\cite{wang2025physctrl} & \underline{22.47} & \underline{0.875} & \underline{0.282} & \underline{100.40} & \underline{83.25} \\
        \midrule
        Ours  & \textbf{25.32} & \textbf{0.895} & \textbf{0.241} & \textbf{90.56} & \textbf{69.54} \\
        \bottomrule
    \end{tabular*}
\end{table}

\noindent\textbf{Generalization beyond simulation}.
Figure~\ref{fig3} presents results on real-world images, which differ substantially in appearance from the synthetic training data. Despite this domain gap, our method predicts plausible motion fields and generates temporally coherent videos. By separating motion modeling from RGB synthesis through an intermediate flow representation, the two-stage design makes the predicted dynamics less dependent on simulator-specific appearance cues.

\subsection{Ablation Studies}
\begin{figure}[t]
    \centering
    \includegraphics[width=1\linewidth]{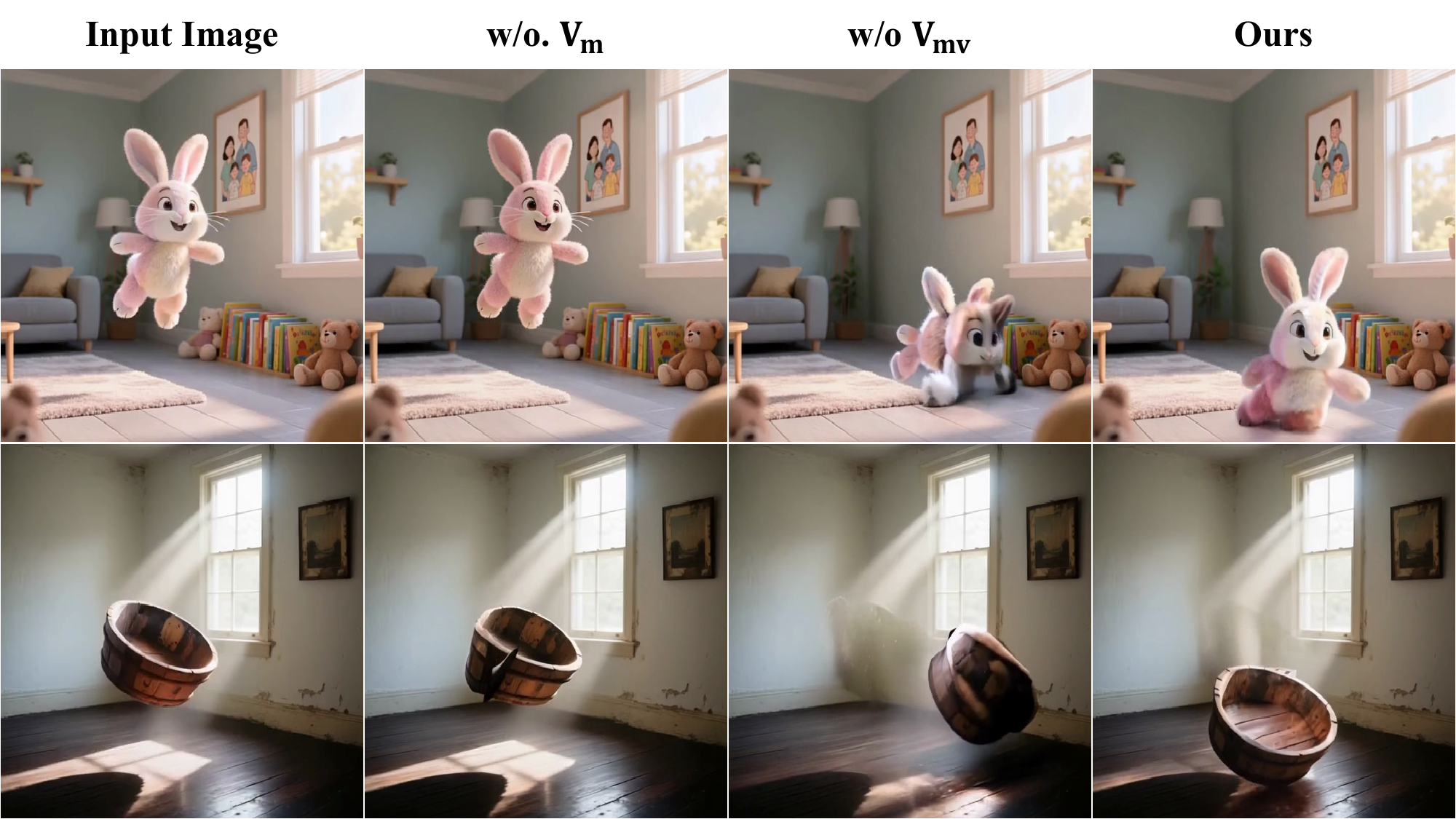}
    \caption{\textbf{Ablation study on conditional videos}. The left column shows the input single image, and the three right columns each display one representative frame from the generated RGB video.}
    \label{fig:ablation_video}
\end{figure}

\noindent\textbf{\textcolor{black}{Ablation study on key modules.}}
\textcolor{black}{PAAM combines a global motion attention module (Motion-Attn) and a local deformation module (Deform-Attn) to model global displacement and local deformation, respectively. Table~\ref{tab:ablation1} reports an incremental ablation starting from the baseline, with SFT, physical conditioning, PAAM, and Phys-FT added in sequence. Each configuration retains everything included in the previous row. Adding PAAM improves all reported metrics, including Physical from 0.594 to 0.608 and PC from 3.29 to 3.37. FlowRender is first trained on OpenVid~\cite{nan2024openvid} and then fine-tuned on WISA~\cite{wang2025wisa}; this Phys-FT stage further raises Physical to 0.613 and PC to 3.41. Figure~\ref{fig:ablation_module} complements the unified table by qualitatively showing the effects of Motion-Attn and Deform-Attn.}

\begin{table}[t]
    \centering
    \caption{\textbf{Incremental ablation of the model, with each row building on the one above.}}
    \label{tab:ablation1}
    \setlength{\tabcolsep}{5.8pt}
    \renewcommand{\arraystretch}{1}
    \begin{tabularx}{\linewidth}{@{}l|cc|c|cccc@{}}
    \toprule
    \textbf{Method} & \textbf{Motion}$\uparrow$ & \textbf{Image}$\uparrow$ & \textbf{Physical}$\uparrow$ & \textbf{SA}$\uparrow$ & \textbf{PC}$\uparrow$ & \textbf{Rule}$\uparrow$  \\
    \midrule
    Baseline & 0.992 & 0.628 & 0.570 & 2.66 & 3.14 & 0.221\\
    +SFT & 0.994 & 0.633 & 0.588 & 2.72 & 3.25 & 0.253 \\
    +Condition & 0.994 & 0.634 & 0.594 & 2.76 & 3.29 & 0.258 \\
    +PAAM & 0.996 & 0.636 & 0.608 & 2.79 & 3.37 & 0.267\\
    +Phys-FT & \textbf{0.997} & \textbf{0.640} & \textbf{0.613} & \textbf{2.81} & \textbf{3.41} & \textbf{0.281} \\
    \bottomrule
    \end{tabularx}
\end{table}

\begin{table}[t]
    \centering
    \caption{\textbf{Ablation on mask and multi-view priors.}}
    \setlength{\tabcolsep}{3.0pt}
    \renewcommand{\arraystretch}{1}
    \begin{tabularx}{\linewidth}{@{}l|cc|c|ccc@{}}
    \toprule
    \textbf{Method} & \textbf{Motion}$\uparrow$ & \textbf{Image}$\uparrow$ & \textbf{Physical}$\uparrow$ & \textbf{SA}$\uparrow$ & \textbf{PC}$\uparrow$ & \textbf{Rule}$\uparrow$ \\
    \midrule
    w/o mask prior & 0.992 & 0.638 & 0.584 & 2.62 & 3.24 & 0.236 \\
    w/o multi-view prior	 & 0.995 & 0.636 & 0.605 & 2.78 & 3.29 & 0.248 \\
    Ours & \textbf{0.997} & \textbf{0.640} & \textbf{0.613} & \textbf{2.81} & \textbf{3.41} & \textbf{0.281} \\
    \bottomrule
    \end{tabularx}
    \vspace{-0.3cm}
    \label{tab4}
\end{table}

\begin{figure}[t]
    \centering
    \includegraphics[width=1\linewidth]{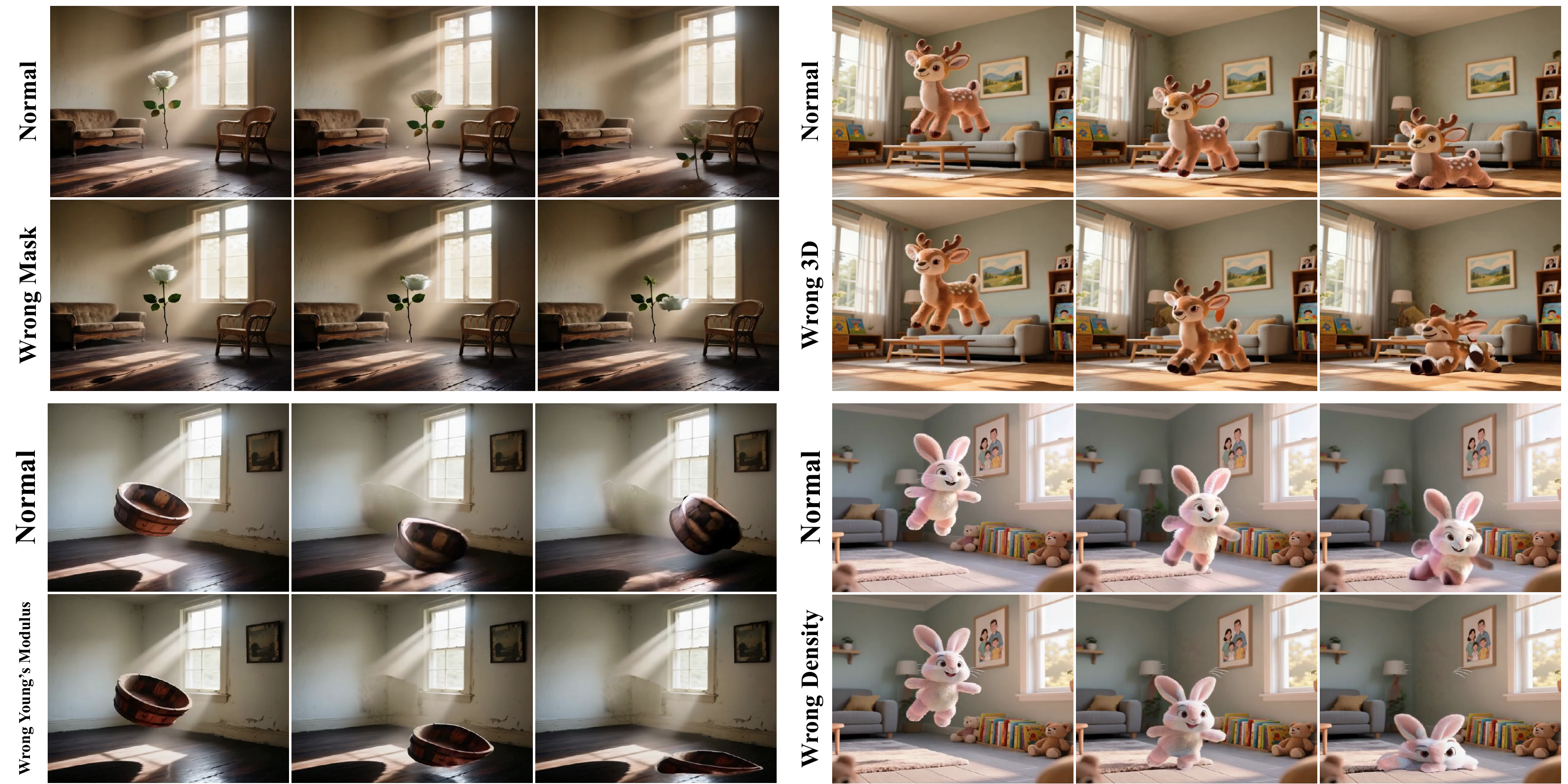}
    \caption{\textbf{Physics-controlled results with abnormal inputs}. With extremely incorrect physical attributes or priors, the generated outputs can be anomalous.}
    \label{fig5}
\end{figure}

\begin{figure*}[t]
    \centering
    \includegraphics[width=1\linewidth]{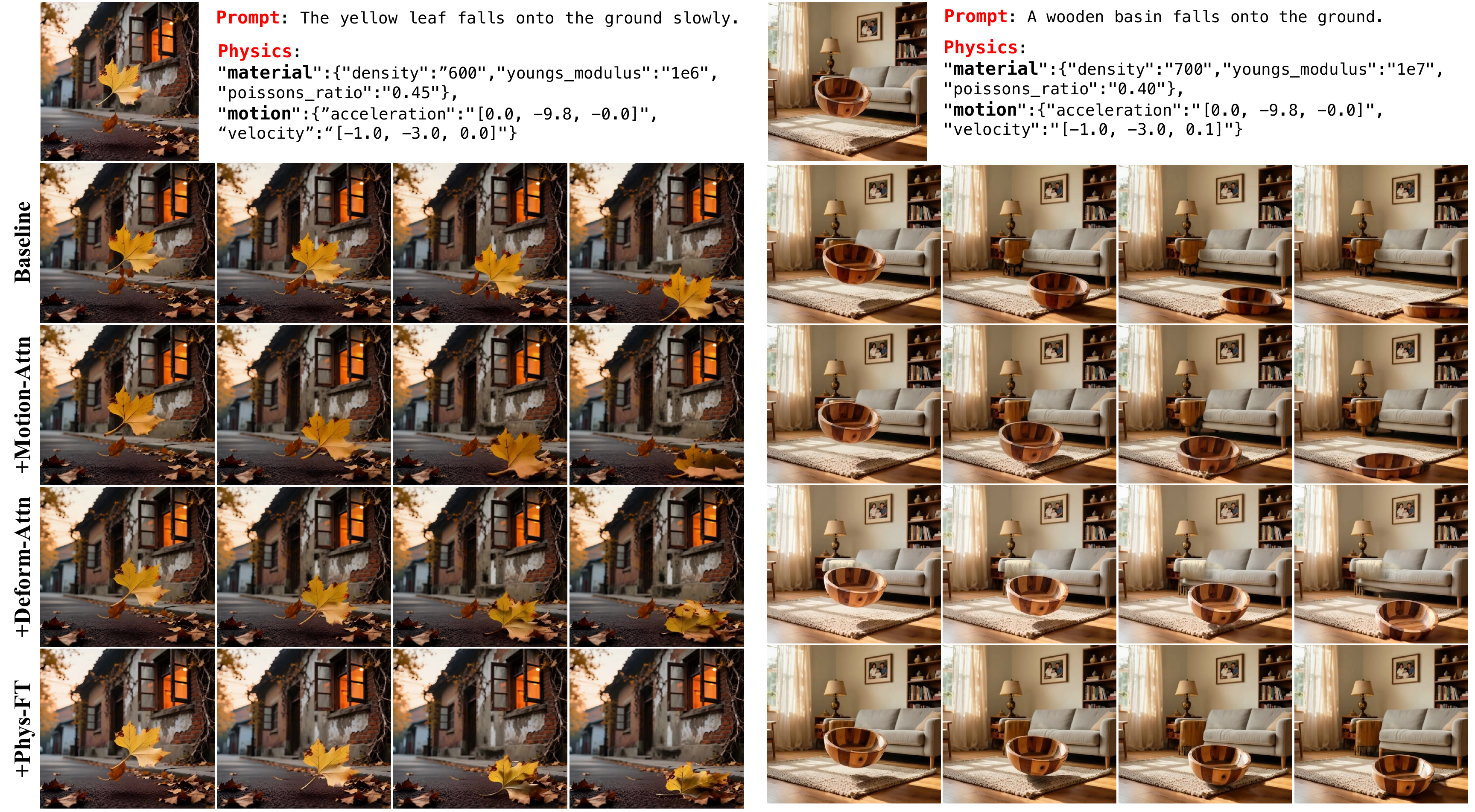}
    \caption{\textbf{Qualitative ablation of key components.} For each example, all variants use the same input image, text prompt, and physical conditions, with frames arranged temporally from left to right. From top to bottom, Motion-Attn, Deform-Attn, and physics-focused fine-tuning (Phys-FT) are progressively added to the baseline, illustrating their complementary contributions to global motion, local deformation, and overall physical plausibility.}
    \label{fig:ablation_module}
\end{figure*}

\begin{table}[t]
    \centering
    \caption{\textbf{Ablation study on proposed loss functions}. The results are evaluated with GPT-4o.}
    \label{tab:ablation2}
    \setlength{\tabcolsep}{2pt}
    \renewcommand{\arraystretch}{1.0}
    \begin{tabular*}{\linewidth}{@{\extracolsep{\fill}} l|cc|c|ccc @{}}
        \toprule
        \textbf{Method} & Motion$\uparrow$ & Image$\uparrow$ & Physical$\uparrow$ & SA$\uparrow$ & PC$\uparrow$ & Rule$\uparrow$ \\
        \midrule
        w/o $\mathcal{L}_{vel}$ & 0.996 & 0.637 &  0.607 & 2.75 & 3.36 & 0.277 \\
        w/o $\mathcal{L}_{acc}$ & 0.997 & 0.638 & 0.610 &  2.78 & 3.38 & 0.271 \\
        w/o $\mathcal{L}_{ela}$ & 0.995 & 0.635 & 0.605 & 2.75 & 3.32 & 0.266\\
        \textbf{Ours} & \textbf{0.997} & \textbf{0.640} & \textbf{0.613} &\textbf{2.81} &\textbf{3.41} &\textbf{0.281} \\
        \bottomrule
    \end{tabular*}
\end{table}

\noindent\textbf{\textcolor{black}{Ablation on priors}.}
To improve PA-Flow’s semantic and 3D consistency, we condition the model on the mask video $V_m$ and multi-view video $V_{mv}$. Figure~\ref{fig:ablation_video} shows that removing $V_m$ leads to ambiguous motion regions, while removing $V_{mv}$ causes unrealistic deformations from weak global 3D awareness.
As shown in Table~\ref{tab4}, removing the mask prior or multi-view prior consistently degrades performance, especially on physical plausibility, reflecting the importance of these components.

\noindent\textbf{\textcolor{black}{Ablation study on loss functions.}}
\textcolor{black}{We also evaluate the effectiveness of the proposed loss functions, including the velocity-smoothness loss $\mathcal{L}_{vel}$, acceleration-consistency loss $\mathcal{L}_{acc}$, and elastic strain energy loss $\mathcal{L}_{ela}$. These losses are designed to provide complementary motion regularization: $\mathcal{L}_{vel}$ penalizes abrupt velocity changes, $\mathcal{L}_{acc}$ penalizes inconsistent acceleration patterns, and $\mathcal{L}_{ela}$ introduces an elastic-energy prior for deformation regions. As shown in Table~\ref{tab:ablation2}, removing any of these losses degrades performance, confirming that both temporal regularity and deformation constraints are useful. Among them, removing $\mathcal{L}_{ela}$ leads to the largest drop, indicating that elastic strain energy plays a particularly important role in modeling non-rigid deformation and improving overall physical plausibility.}

\noindent\textbf{Failure case analysis}. 
Our model supports controllable video generation with respect to both material and motion attributes.
When the specified physical attributes or input priors deviate substantially from plausible values, the generated results may exhibit visually implausible physical behavior. For failure case analysis, Figure~\ref{fig5} presents four representative examples arising from different types of upstream errors. Incorrect mask priors may activate unintended motion regions, inaccurate 3D priors may lead to unrealistic deformation, and inappropriate values of Young's modulus or density may result in excessive deformation.

\section{Conclusion}
\label{sec:con}
\textcolor{black}{In this work, we presented a physics-aware video generation framework for object dynamics spanning rigid-body motion, elastic deformation, and flexible-body motion. The framework disentangles object-motion generation from motion-guided texture synthesis.}
Our two-stage design supports physics-aware optical flow generation and optical flow-guided plausible video generation without relying on physics engines. 
Furthermore, we provide a large-scale physics-engine–based video dataset that uniquely combines realistic backgrounds with detailed annotations of material and motion properties.
Experimental results demonstrate that our method achieves superior motion consistency and visual quality compared with prior methods.

\vspace{0.5cm}

 
%
\bibliographystyle{IEEEtran}
\bibliography{main}


\end{document}